\pdfoutput=1

\documentclass[11pt]{article}

\usepackage{emnlp2021}

\usepackage{times}
\usepackage{latexsym}

\usepackage[T1]{fontenc}

\usepackage[utf8]{inputenc}

\usepackage{microtype}
\usepackage{graphicx}
\usepackage{multirow}
\usepackage{pifont}
\usepackage{lipsum}
\usepackage{float}
\usepackage{booktabs}
\usepackage{color,colortbl}

\newcommand{\cmark}{\ding{51}}%
\newcommand{\xmark}{\ding{55}}%
\definecolor{LightRed}{rgb}{1,0.88,1}
\definecolor{LightBlue}{rgb}{0.88,1,1}
\definecolor{LightYellow}{rgb}{1,1,0.88}

\title{Cleaning Dirty Books:  Post-OCR Processing for Previously Scanned Texts}

\author{Allen Kim, Charuta Pethe, Naoya Inoue, Steven Skiena \\
  Department of Computer Science, \\ Stony Brook University, NY, USA \\
  {\texttt{\{allekim,cpethe,ninoue,skiena\}@cs.stonybrook.edu}}}

\begin{document}
\maketitle
\begin{abstract}
Substantial amounts of work are required to clean large collections of digitized books for NLP analysis, both because of the presence of errors in the scanned text and the presence of duplicate volumes in the corpora.
In this paper, we consider the issue of deduplication in the presence of optical character recognition (OCR) errors.  We present methods to handle these errors, evaluated on a collection of 19,347 texts from the Project Gutenberg dataset and 96,635 texts from the HathiTrust Library.
We demonstrate that improvements in language models now enable the detection and correction of OCR errors without consideration of the scanning image itself.
The inconsistencies found by aligning pairs of scans of the same underlying work provides training data to build models for detecting and correcting errors.
We identify the canonical version for each of 17,136 repeatedly-scanned books from 58,808 scans.  Finally, we investigate methods to detect and correct errors in single-copy texts. We show that on average, our method corrects over six times as many errors as it introduces. We also provide interesting analysis on the relation between scanning quality and other factors such as location and publication year.
\end{abstract}

\section{Introduction}
\label{sec:introduction}
The HathiTrust and Gutenberg corpora are critical resources for literary analysis and NLP research, providing legal access to tens of thousands of texts for research purposes.

Both were constructed from scanned texts, with manual correction in the case of the Gutenberg corpus. These efforts for the Gutenberg Project have begun as early as the 1970s when there was a foreseeable need to digitize open-domain books. Furthermore, the HathiTrust dataset also was constructed from a compilation of books from multiple libraries from universities and states, and adds a substantial amount of extra content.

However, when compiling a library of books from multiple sources, many challenges arise in maintaining a well-structured catalog with minimal redundant data. Quite often, a popular book will appear in multiple sources of differing quality.

In this paper, we describe a major effort to clean and organize these texts to provide a stronger foundation for NLP research in literary texts. Our main contributions are:

\begin{itemize}
    \item \emph{OCR correction of previously scanned texts} -- Book scanning technologies are a mix of vision and language analysis, with language models used to correct the visual processing errors and ambiguity inherent in the scanning process.   Language models are now substantially more powerful than available at the time the bulk of the Gutenberg/HathiTrust corpora were collected.   
    We employ these language models for detecting and correcting scanning errors, yielding much cleaner texts for downstream analysis. These cleaned texts will be made available to the research community subject to the limits enforced by Project Gutenberg and HathiTrust.
    
    With these models, we find errors in hundreds of Gutenberg books. Some examples are shown in Table \ref{tab:guten_ocr}.
    
    \begin{table}[]
        \centering
        \begin{tabular}{c|c}
        Gutenberg ID & Incorrect Sentence \\ \hline
        3005 & He returned \textbf{hone}\\
        5798 & I \textbf{dod} not smoke. \\
        12773 & Which would \textbf{he} absurd \\
        44223 & ...pleaded \textbf{tie} major \\
        53604 & What did he \textbf{clo}? 
        \end{tabular}
        \caption{Examples of errors detected in Project Gutenberg books by our method}
        \label{tab:guten_ocr}
    \end{table}
    
    We do note that it may not be so clear cut at times due to intentional misspellings in dialogue. For example, "Tat will pe wrong" is a legitimate sentence (dialogue) in Malcolm by George MacDonald, but is detected as an error. In general, we find approximately 18.9\% of our detected errors in HathiTrust books to be within quotes, and we see that books with unusually high error rates in quotes generally stem from OCR errors found on question marks ('P' instead of '?') or books with heavy vernacular English such as ``On the Plantation'' by Joel Chandler Harris (ex. ``kaze I'm dat ole dat I ain't '').
    
    We show that on average, our model fixes more than six times as many errors as it introduces. Even among the ``errors'' the model introduces, much of them may actually improve downstream NLP tasks even if the new words may be against the author's original intentions. This is particularly true in books with heavy accented English.
    
    \item \emph{Alignment analysis of repeatedly-scanned books} -- We leverage the presence of several thousand books which have been scanned two or more times across the union of the Gutenberg/HathiTrust, permitting us to pinpoint exactly where differences exist in each pair of texts. By employing the language-based OCR correction models described above, we can identify the correct variant of the text with high confidence, providing training data to improve correction models.  Our alignment procedure permits us to identify the better of the two versions and construct a single canonical text of higher quality than either of the input source texts -- as well as train models to clean up singleton texts.
    
    We collect 8,430,587 aligned differences, which were split into a training and test dataset for our models and provide them for public use, again subject to limits enforced by Gutenberg and HathiTrust.
    
    \item \emph{Analysis of scanning errors} -- Our alignment methodology provides detailed information about the causes of observed scanning errors in the HathiTrust corpus. We identify defect levels as a function of library/location, publication date, and character signatures. We show one such result in Table \ref{tab:libquality}, which shows the quality of a subset of books from different libraries. In general, we find that location is not as big of a factor as the source that digitized their books, primarily Google versus the Internet Archive. These results shed interesting light on the history of printing, and serve to create prior distributions for improved scanning technologies.

\begin{table}[t]
    \centering
    \begin{tabular}{c|c|c|c|c}
    ID & Location & Count & Year & Quality \\ \hline
    \rowcolor{LightBlue}
    nyp & NYPL & 2071 & 1903 & 0.879 \\
    \rowcolor{LightYellow}
    miun & Univ. MI & 6 & 1905 & 0.879 \\
    \rowcolor{LightBlue}
    mdp & Univ. MI & 1740 & 1904 & 0.866 \\
    \rowcolor{LightBlue}
    nnc1 & Columbia & 44 & 1893 & 0.852 \\
    \rowcolor{LightBlue}
    uva & Univ. VA & 143 & 1904 & 0.847 \\
    \rowcolor{LightBlue}
    pst & PSU & 24 & 1895 & 0.844 \\
    \rowcolor{LightBlue}
    njp & Princeton & 437 & 1893 & 0.841 \\
    \rowcolor{LightBlue}
    uc1 & UC & 446 & 1898 & 0.837 \\
    \rowcolor{LightBlue}
    wu & Univ. WI & 78 & 1999 & 0.824 \\
    \rowcolor{LightBlue}
    inu & Univ. IN & 98 & 1897 & 0.819 \\
    \rowcolor{LightBlue}
    coo & Cornell & 36 & 1905 & 0.773 \\
    \rowcolor{LightBlue}
    umn & Univ. MN & 12 & 1904 & 0.762 \\
    \rowcolor{LightBlue}
    ien & NW Univ. & 2 & 1920 & 0.733 \\
    \rowcolor{LightRed}
    nc01 & UNC & 118 & 1894 & 0.715 \\
    \rowcolor{LightRed}
    uc2 & UC & 1290 & 1901 & 0.705 \\
    \rowcolor{LightRed}
    uiuo & Univ. IL & 42 & 1883 & 0.694 \\
    \rowcolor{LightRed}
    loc & Congress & 46 & 1901 & 0.669 \\
    \rowcolor{LightRed}
    dul1 & Duke & 43 & 1891 & 0.649 \\
    \rowcolor{LightBlue}
    hvd & Harvard & 23 & \textbf{1832} & 0.566 \\
    \end{tabular}
    \caption{Quality of sampled books by location - blue means the books were digitized by Google, red means the books were digitized by the Internet Archive, yellow means the books were locally digitized (at the location specified). The `Year' column shows the average publication year, which explains the lower quality for books scanned by Harvard, since these are significantly older.}
    \label{tab:libquality}
\end{table}
    
\end{itemize}

\section{Background}

\subsection{Project Gutenberg}
Project Gutenberg is one of the oldest online libraries of free eBooks that currently has more than 60,000 available texts \cite{project_gutenberg}. Given the wide range of languages and topics available, we restrict ourselves to English fiction, which narrows the scope of text to about 19,347 books. For each book, in addition to the text, we are given the title, author, and subject as metadata.

\subsection{HathiTrust}
The HathiTrust digital library is a collaborative effort between academic and research libraries to provide a unified corpus of books that currently number over 8 million book titles \cite{hathitrust_dataset}. By filtering down to English fiction books in this dataset using provided metadata \cite{underwood_dataset}, we get 96,635 books along with extensive metadata including title, author, and publishing date. 

\subsection{Related Work}

\paragraph{OCR post-analysis.} OCR post-detection and correction has been discussed extensively and can date back before 2000, when statistical models were applied for OCR correction \cite{kukich1992techniques,tong1996statistical}. These statistical and lexical methods were dominant for many years, where people used a combination of approaches such as statistical machine translation with variants of spell checking \cite{bassil2012ocr,evershed2014correcting,afli2016ocr,kissos2016ocr,schulz2017multi,coustaty2018adaptive}. These approaches were also combined with a human aspect, where an interface could be presented to a human corrector that provide aligned text. A human corrector can then efficiently correct mistakes in bulk \cite{taghva2001ocrspell,vobl2014pocoto}.

We also make note of other data cleaning models that have relied on automatas or generative models \cite{kolak2003generative,pasula2003identity,mayfield2009statistical,llobet2010ocr,abedjan2016detecting,lew2021pclean}. Methods such as PClean work off of Bayesian principles and probabilistic programming to identify likely errors in a specific domain.

In addition to these models, there have been analysis and visualizations on the OCR errors themselves on digital libraries \cite{chiron2017impact}. \newcite{jatowt2019deep} show interesting statistical analysis of OCR errors such as most frequent replacements and errors based on token length over several corpora . These provide insight into the most common sources of errors and also show how different sets of documents each present their own individual features. It is shown that one cannot generalize assumptions about OCR to all domains.

\paragraph{ICDAR Competitions.} With growing interest in these fields, the ICDAR Competition on Post-OCR Text Correction was hosted during both 2017 and 2019 \cite{chiron2017icdar2017,rigaud2019icdar}. These competitions called for participants to submit their best models for both OCR detection and correction with a provided training dataset that aligned dirty text with ground truth. The difference in the models submitted between these two years highlight the advancements in natural language processing. 

In ICDAR 2017, the top OCR correction models focused on neural methods. Neural machine translation had been shown to outperform statistical machine translation on many tasks, and the top team's approach explored both these models and combined results from multiple sources \cite{amrhein2018supervised}. In the 2019 competition, the best performing team was CCC, using BERT for fine tuning and character-level machine translation for error correction. Many others have began to build off of this same structure. For example, \newcite{nguyen2020neural} present post-OCR approaches based on a contextual language model (BERT) and neural machine translation (NMT) on aligned text, as done by CCC. They improve upon them by applying static word embeddings to improve error detection, and applying length difference heuristics to improve correction output.

\paragraph{Vernacular English.} Another related direction connected to OCR errors is analysis of text with vernacular English. In general, different dialects in English do not affect understanding for native English speakers as much as they affect current NLP systems. This has been considered by Tan et al. \cite{tan2020mind}, proposing a new encoding scheme for word tokenization to better capture these variants. One can also consider applying OCR correction models that work at a token level to normalize such texts into proper English as well.

\paragraph{Language Models.} Separate from OCR errors, we also make use of concepts in language models. Language models have provided a means to evaluate the likelihood of various phrases. Traditionally, this was done with n-gram models \cite{bengio2003neural}, but this has been replaced with neural language models. With the advent of transformers in the form of BERT and RoBERTa, language models have progressed even further \cite{devlin2018bert, liu2019roberta}. In recent years, masked language model scoring illustrates a way make use of the transformer architectures to provide scoring of sentences \cite{salazar2019masked}. There have also been advances in deeper models such as GPT2 that provide even stronger results as well \cite{radford2019language}.

\section{Alignment Methods}
\label{sec:dedup}
We focus on a collection of books from the HathiTrust dataset of which we have 96,635. Our first task was to find duplicate books and to align the content such that we could find the differences.
\subsection{Deduplication}
Given a large collection of text, we first identify which texts should be grouped together as a ``deduplicated'' set. We refer to a deduplicated set of books as a set of texts in which each text corresponds to the same overall content. There may be variations in the content due to editing or OCR differences, but the majority of the text should be similar.

To check for similarity, we use the contents of the books with the n-gram overlap as a metric. In our case, we process the texts into a set of five-grams and impose at least a 50\% overlap between two sets of five-grams for them to be considered the same. In practice, duplicate books have an overlap ratio close to 100\%, and different books have overlap ratios close to 0\%, so the 50\% threshold is insensitive to small changes.

One can consider checking similarity between book titles and authors as a way to deduplicate books, but this is not a practical approach. Titles of the same book can vary with different editions; thus, fuzzy matching becomes a necessity. However, it becomes unclear at what threshold one should consider it a match. If it is too strict, books that should be clustered might be missed while if it is too loose, then there may be too many false positives between books of similar titles. There may also exist annotation errors in the metadata as well, which requires looking into the actual content of the book.

To avoid comparing each text to every other text, which would be quadratic in the corpus size, we first group books by author and compute the pairwise overlap score between each book in each author group. To then deduplicate the sets, we treat the problem as finding the connected components in a graph, where the nodes are books and edges exist between books that were found to be similar.

\paragraph{Anthologies} There is one issue regarding books that contain the contents of many other books (anthologies). We first filter these books out to avoid situations that break transitivity. For example, if book A includes book B and book C in its contents, we would get that book A is similar to book B and book C, but book B and C may not be similar to each other. Thus, to differentiate between anthologies and books that are legitimate duplicates, we consider the titles and lengths of the books in common. If there are no common tokens among the titles and the parent book is longer than the others, we consider the parent book an anthology. We also filter out books that are of the form "Works, Works of ..., The complete writings of ..., The novels of ..." and related variants. In total, we find 11,382 anthologies out of our HathiTrust dataset of 96,634 books and 106 anthologies from our Gutenberg dataset of 19,347 books.

\subsection{Text Alignment}

Given the set of deduplicated books, our task is to now align the text between books. More concretely, the task is: given two tokenized books of similar text (high n-gram overlap), create an alignment between the tokens of both books such that the alignment preserves order and is maximized. At its core, this problem is simply a longest common subsequence problem done at a token level. We show an example of such an alignment in Table \ref{tab:textalign}. The only problem is that the running time of the dynamic programming solution is proportional to product of the token lengths of both books, which is too slow in practice.

\begin{table}
    \centering
    \begin{tabular}{c|c|c|c|c|c|c}
    I & \textbf{kndr} & \textbf{ft} & it & isn't & my & business  \\\hline
    I & \textbf{know} & & it & isn't & my & business    
    \end{tabular}
    \caption{Example of text alignment - the words "I" and "it" are aligned and the bold words between them are the differences.}
    \label{tab:textalign}
\end{table}

\begin{table*}
    \centering
    \begin{tabular}{p{3.75cm}|p{3.75cm}|c|c|c|c}
    Correct & Incorrect & Baseline & BERT & RoBERTa & GPT2 \\ \hline
     ...had no doubt & ...has no donbt & \cmark & \cmark & \cmark & \cmark \\
     ...I had laid my head & ...I bad laid my head & \xmark & \cmark & \cmark & \cmark \\
     ...clinging flakes of froth & ...clinging takes of froth & \xmark & \xmark & \cmark & \cmark \\
     ...the senor & ...the sefior & \xmark & \xmark & \xmark & \cmark \\ \hline
     \multicolumn{2}{c|}{Accuracy of models over 1000 human annotated pairs:} & 0.712 & 0.761 & 0.794 & 0.853
    \end{tabular}
    \caption{Examples of extracted pairs with results from different models with accuracy scores over 1000 pairs(the annotators and models judge the phrase in the context of the full sentence).}
    \label{tab:model comparisons}
\end{table*}

To remedy this problem, we employ the use of ``anchor'' tokens, which are tokens that occur only once in a book. Some examples of such tokens are the words ``systematic'', ``rampacious'', ``affix'' in Oliver Twist. They are singleton words that tend to be more specific in meaning. For an average-length book, there only exist a few thousand of these tokens, and thus, we can first align the book according to these tokens. Since the contents of the books are similar, the anchor tokens for both books should also be similar. Thus, we run the full dynamic programming solution between the anchor tokens of both books, which can be done much faster than the book in its entirety. Once we have the alignment between the anchor tokens, we can then run the dynamic program between each aligned anchor token. In general, these distances are quite short and thus, the overall running time improves dramatically. Note that anchor n-grams would also work if there is not enough anchor tokens.

\subsection{Rating Sentence Pairs}

Given the alignment between a pair of books, we now identify where the differences lie. For each consecutive aligned token, we check whether there is a gap in alignment in either of the books. At every point where a gap lies, we capture those areas as token-wise differences as well as the sentences in which these differences lie.

The main question now is: given two similar sentences with some small difference between them, which sentence is ``more'' correct? Generally, these differences can be attributed to OCR errors, typically random letters or punctuation appearing in text. Other times, it may be errors where letters are replaced such as `m' by `in' or `2' by `?'.

\paragraph{Baseline.} We first consider a baseline of a dictionary lookup. Given a sentence, we consider the ratio of tokens that are in a dictionary \footnote{We use the NLTK English dictionary.} to the total number of tokens in the sentence . We consider the sentence that has a higher ratio to be the better sentence; if equal, we select randomly.

However, this is quite often not sufficient as the ratio tends to be roughly equal for both sentences. This can be attributed to the differences in both sentences being out of dictionary, such as when a name gets misspelled or both being in the dictionary such as when both are legitimate words (ex. `but' versus `nut' as errors). Additionally, there may be multiple errors in the same sentence, resulting in skewed ratios. Also, sentences may not always be of the same length due to OCR errors among sentence-defining punctuation such as periods. Due to these factors, we turn towards stronger models.

\paragraph{Language Models.} Thus, we rely on language models, particularly models based on modern transformer architectures. In this context, we can apply language models to compute the likelihood of a given sentence based on the probability of each token within it. For a given sentence, we compute its likelihood by passing it through a given language model and compute the log sum of token probabilities normalized by the number of tokens, to avoid biasing on sentence length. Thus, given two sentences, we can compute the normalized log likelihood for both and choose a winner based on the greater value.

\paragraph{Evaluation.} For our experiments, we test the baseline along with three language models based on BERT \cite{devlin2018bert}, RoBERTa \cite{liu2019roberta}, and GPT2 \cite{radford2019language}. For all of these models, we use the pretrained models without any fine tuning. For the test set, we procure a random set of 1000 pairs of sentences from our corpus, and manually annotate which sentence is better for each one. We note that there are 93 pairs that were deemed ambiguous by the human annotators; thus, they were not included in the final evaluation. Table \ref{tab:model comparisons} shows the results for this human annotated set with some examples.

\paragraph{Analysis.} While the baseline performed respectably compared to random guessing (0.5), we find that GPT2 performs the best out of all the methods.   Thus, we apply GPT2 as the main language model for determining the correct sentence. We do note that it is possible for both sentences to contain errors, but we can still apply the same methodology to judge which of the two is less severe.

\subsection{Determining Best Books}

Given a pair of duplicate books, we consider the task of identifying the one that is of better quality from an OCR perspective. By applying the text alignment and sentence evaluations described in the prior subsections, we compute a list of aligned sentence pairs between the two books with the likelihood scores for each one. We can convert these scores into a confidence by normalizing with softmax. 

Formally, given two books $B_1$ and $B_2$ with $n$ aligned sentences, we consider $p_i, q_i$ as the respective confidence scores for the $i^{th}$ aligned sentence pair ($p_i + q_i = 1$).

The simplest method to determine the better of the two books then would be to take the majority count. Whichever book is favored more among all the sentence pairs can be considered the winner. Concretely, we compute the probability that book 1 is better than book 2 as:
\begin{equation}
\label{eq:one}
\Pr[B_1 > B_2] = \frac{\texttt{count}(p_i > q_i)}{n}
\end{equation}  
If this is greater than 0.5, $B_1$ is declared the winner; otherwise, $B_2$ is the winner. Often, this works well but when the number of errors are relatively balanced between both books, then we need to consider the confidence scores themselves.

To address this issue, we apply a Bayesian updating approach. Recall that the posterior probabilities are proportional to the product of the likelihood and prior. As the prior, we use Equation \ref{eq:one}, and we compute the log posterior for $B_1$ ($B_2$ is analogous) as:
\begin{equation}
\underbrace{\sum_{i=1}^n \log{p_i}}_{\textrm{log likelihood}} + \underbrace{\log{\left(\frac{\texttt{count}(p_i > q_i)}{n}\right)}}_{\textrm{log prior}}
\end{equation}

The final winner is decided by comparing the final two log posteriors, and choosing the book corresponding to the larger value.

So far, we have only discussed comparisons between two given books. However, a general set of duplicates may contain more than two books. To find a winner among an arbitrary sized set of books, we employ a tournament strategy. We use our Bayesian approach to find the winner between distinct pairs of books, and the winner of each pair face off, and so on until there is only one winner. It is the final winner of the tournament that is marked as the canonical text of the set. 

We apply our method on the full 96,635 HathiTrust texts, and find 58,808 of them to be a duplicate to another book in the set. Among the duplicates, we identify 17,136 canonical books.

\paragraph{Golden Dataset.} To evaluate our approach, we create a golden dataset based on an alignment between Gutenberg and HathiTrust. By applying the same deduplication methods discussed in Section \ref{sec:dedup}, we create a test dataset of 6,694 paired books. In this set, we use the Gutenberg version as the ground truth since Gutenberg books are of higher quality due to human editors compared to HathiTrust books. To evaluate our method for choosing a canonical book, we apply it on our golden dataset to see how often it selects Gutenberg over HathiTrust as the better copy. We find that it selects the Gutenberg version 6,059 times out of the total 6,694 books, showing that our method preferred Gutenberg 90.5\% of the time. This agrees with our understanding that Gutenberg books are of higher quality.

\section{OCR Errors in Single-Copy Texts}

We now consider OCR errors for single copy texts. In this setting, we cannot use any alignment technique as the books live in isolation. For this case, we train models for both OCR error detection and correction using the 17,136 sets of duplicate books and their alignments. All models were run on a compute server with 2.30 GHz CPU and TeslaV100 GPU. No hyperparameter tuning was done on any models; default values were run for all models. 

\begin{table*}[htbp]
    \centering
    \begin{tabular}{p{6cm}|p{2.5cm}|p{2.5cm}|c|c}
    \textbf{Sentence} & \textbf{Ground Truth} & \textbf{Generated} & \textbf{Score} & \textbf{Precision} \\ \hline
    ... speaking as gently , as if he had been lying in a satin \textbf{<ocr> eradle </ocr>} . & cradle & robe & 0.263 & 0.776 \\ \hline
    ... know of your brother 's \textbf{<ocr> apphcation n </ocr>} to me ? ''	& application & approbation & 0.398 & 0.829 \\ \hline
    ... was to \textbf{<ocr> seck </ocr>} a home with some friends ... & seek & find & 0.512 & 0.868 \\ \hline
    ... and \textbf{<ocr> dryexposition n </ocr>} of the glories of the house ... & dry exposition & dry exposition & 0.658 & 0.905 \\ \hline
    ... in finding \textbf{<ocr> tlie </ocr>} auger holes . & the & the & 0.998 & 0.992
    \end{tabular}
    \caption{Examples of generated OCR corrections - score represents the confidence in the generated text and precision is calculated across the test set with the corresponding score as a threshold}
    \label{tab:example_generated}
\end{table*}

\subsection{Detecting OCR Errors}

For OCR detection, we want to be able to identify which tokens in a given text can be marked as an OCR error. This is a classic token classification problem; thus, we train RoBERTa-large with a token classification head for 3 epochs.

\begin{figure}[H]
    \centering
    \includegraphics[width=0.8\linewidth]{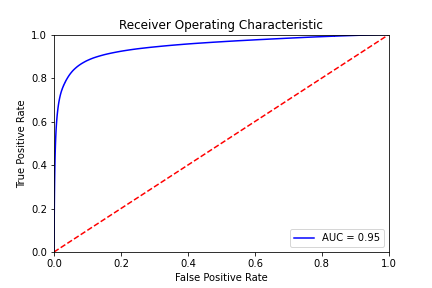}
    \small
    \begin{tabular}{c|cc}
    \textbf{Threshold} & \textbf{Precision} & \textbf{Recall} \\ \hline
    0.25 & 0.699 & 0.701\\
    0.5 & 0.787 & 0.618\\
    0.75 & 0.853 & 0.506\\
    0.95 & 0.916 & 0.234
    \end{tabular}
    \caption{ROC and metrics for OCR detection at various thresholds (in general, we value precision over recall)}
    \label{fig:roc_ocr}
\end{figure}

The training data is derived from our aligned books from before. For each sentence pair, we choose the lower-scoring sentence as the sentence with the OCR error and annotate the tokens as either 0 or 1, where 1 represents an error. We note that tokenization in RoBERTa further breaks down the tokens to sub-tokens. In cases where the word that is marked with an OCR error is broken down into sub-tokens, we label each sub-token as an error. 

We perform a train-test split at the book level, and sample a training set of 2,080,328 sentences, half of which have no OCR errors and half of which do.

Figure \ref{fig:roc_ocr} shows the ROC curve and metrics on the test set. We find that with a high enough threshold, we can opt for a high precision with relatively few mistakes. If the goal is to improve the quality of a book, we prefer to optimize precision over recall as it is more important to be confident in the changes one makes as opposed to trying to catch all of the errors in a book. Empirically, we found a threshold of 0.95 to provide a good balance between prioritizing precision while finding a non-trivial number of errors.

\subsection{Correcting OCR Errors}

For OCR correction, we now assume we have the output of our detection model, and we now want to generate what the correct phrase should be. We model this as a sequence-to-sequence problem, where the input is a sentence containing an OCR error and the output is what the corrected form should be. To do this, we train a base-T5 seq2seq model \cite{raffel2019exploring} with a language modeling head for conditional generation, for 3 epochs.

We use special \textsc{<ocr>} and \textsc{</ocr>} tags to denote the start and end of the OCR error location within a sentence respectively. For generation, we use greedy search decoding to generate the most likely sequence of tokens.

We train this model over the same dataset as OCR detection. We note that our training is performed only on text with errors, annotated with the special \textsc{<ocr>} tokens. We also score the generated text from a 0 to 1 scale. To do this, we simply take the minimum probability across the sequence of generated tokens. 

\paragraph{Analysis.} Table \ref{tab:example_generated} shows examples of generated OCR predictions along with their score. We now consider thresholds above which we accept the generated text. The precision is calculated across the entire test set with the corresponding score in its row as a threshold. Note that precision increases with higher thresholds. Empirically, we choose a threshold of 0.95.

One key point to note is that traditionally, many OCR correction models have been character-based, but with recent advances in transfer learning, we find that recent token-based models have significant advantages in terms of memory as well as performance. With access to more context, token-based models have the advantage that they can make sensible predictions that work as synonyms even if the edit distance from the original text may be far. This may not be completely desirable in certain situations where the original words used need to be preserved (e.g. analyzing an author’s vocabulary), but in many cases, this may actually be beneficial for NLP analysis/downstream tasks. Quantifying the improvement on several downstream tasks will be an interesting extension to consider. We do note that when the model suggests replacements that are semantically similar (e.g. “seek” to “find”), but not structurally (e.g. “tlie” to “the”), then it tends to have lower confidence scores.

\begin{figure}[ht]
    \centering
    \includegraphics[width=7.5cm]{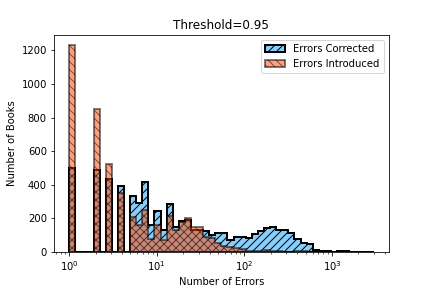}
    \includegraphics[width=7.5cm]{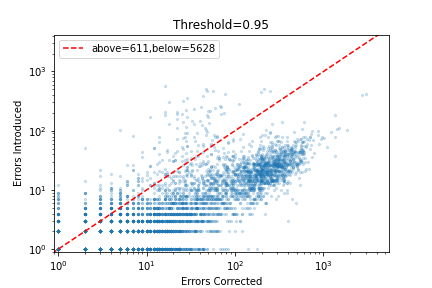}
    \caption{The top figure shows a histogram for the number of errors corrected/introduced. The red bars show how many errors we introduce and the blue bars show how many we correct. The third dark color represents an overlap of the red and blue bars. The bottom figure shows a scatter plot, where each point represents a book. The red line is the identity; thus, points below show books with more fixes than introduced errors, and points above show books with more errors than fixes.}
    \label{fig:detection_to_correction}
\end{figure}

Figure \ref{fig:detection_to_correction} shows the results of OCR correction on our golden test set. In general, we show that we introduce far fewer errors in many books (red bars tend to be more clustered towards the y-axis) compared to how many we correct (blue bars are spread towards the right, indicating that we are correcting many errors for many books). We find that on average, we correct more than six times as many errors as we introduce -- about 61.3 OCR error instances corrected compared to an average 9.6 error instances we introduce. We remark that this is a pessimistic metric as we are only rewarding ourselves for an exact match to our silver-standard ``ground truth'' based on our ranked sentence pairs.

\section{Analysis of Scanning Quality}

\begin{figure}[h]
    \centering
    \includegraphics[width=\linewidth]{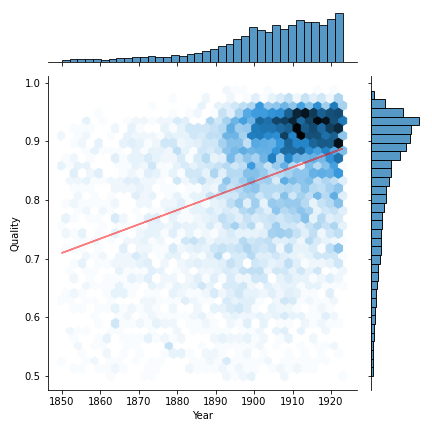}
    \caption{Publication year of books versus quality}
    \label{fig:year_vs_quality}
\end{figure}

\paragraph{Quality by Location.}  The HathiTrust library is a collection of books from multiple sources, mostly composed of universities. We explore whether there are differences in the quality of books depending on location. From Section \ref{sec:introduction}, Table \ref{tab:libquality} shows the quality of books on a subset of books from different libraries. We define the quality of a book to be the percentage of sentences out of the total that do not contain any OCR error. From a presentation on HathiTrust Data in Detail\footnote{\url{https://www.hathitrust.org/documents/HTRC-UnCamp2012-York-201211.pdf}}, we find that some of the books were digitized by Google, others were digitized by the Internet Archive, and few were digitized locally. Overall, we find that the quality of books digitized by Google were of higher quality than the Internet Archive. The exception is Harvard, but this is due to the fact that their books, on average, were published much earlier than the rest of the corpus, and consequently, are of lower quality.

\paragraph{Quality by Publication Year.} We also look at the quality of scans by publication year. Figure \ref{fig:year_vs_quality} shows this relation. In general, we see that quality has improved over the years with many books being of high quality in the early 1900s. Prior to that point, the quality of books was spread out more uniformly.

\paragraph{Top Character Replacements.} Finally, we look at common substitutions for characters. Figure \ref{tab:char_occurence} show the top 3 replacements for each character. Each cell is color-coded by a normalized frequency across all substitutions. We see that some of the most common OCR errors are `j' with `;' and `l' with `!'.

\begin{table}
    \centering
    \begin{tabular}{c|ccc|c|ccc}
Char & 1 & 2 & 3 & Char & 1 & 2 & 3 \\ \hline
a & \cellcolor{green!8} the & \cellcolor{green!6} `` & \cellcolor{green!2} an & n & \cellcolor{green!13} -- & \cellcolor{green!9} a & \cellcolor{green!5} ' \\
b & \cellcolor{green!3} by & \cellcolor{green!1} be & \cellcolor{green!0} is & o & \cellcolor{green!7} of & \cellcolor{green!5} ? & \cellcolor{green!3} to \\
c & \cellcolor{green!7} ' & \cellcolor{green!6} `` & \cellcolor{green!0} * & p & \cellcolor{green!0} P & \cellcolor{green!0} -- & \cellcolor{green!0} ? \\
d & \cellcolor{green!12} -- & \cellcolor{green!3} a & \cellcolor{green!1} , & q & \cellcolor{green!0} a & \cellcolor{green!0} o & \cellcolor{green!0} ? \\
e & \cellcolor{green!17} -- & \cellcolor{green!3} he & \cellcolor{green!2} , & r & \cellcolor{green!7} ? & \cellcolor{green!6} -- & \cellcolor{green!4} , \\
f & \cellcolor{green!47} ? & \cellcolor{green!6} l & \cellcolor{green!5} ! & s & \cellcolor{green!20} -- & \cellcolor{green!11} ! & \cellcolor{green!3} , \\
g & \cellcolor{green!9} -- & \cellcolor{green!1} , & \cellcolor{green!0} ; & t & \cellcolor{green!15} ? & \cellcolor{green!8} -- & \cellcolor{green!6} ! \\
h & \cellcolor{green!1} he & \cellcolor{green!1} a & \cellcolor{green!1} it & u & \cellcolor{green!78} `` & \cellcolor{green!6} a & \cellcolor{green!1} up \\
i & \cellcolor{green!18} l & \cellcolor{green!8} ? & \cellcolor{green!7} 1 & v & \cellcolor{green!1} " & \cellcolor{green!0} , & \cellcolor{green!0} `` \\
j & \cellcolor{green!80} ; & \cellcolor{green!3} , & \cellcolor{green!1} ' & w & \cellcolor{green!5} `` & \cellcolor{green!2} " & \cellcolor{green!1} -- \\
k & \cellcolor{green!2} it & \cellcolor{green!0} a & \cellcolor{green!0} -- & x & \cellcolor{green!1} " & \cellcolor{green!0} 1 & \cellcolor{green!0} X \\
l & \cellcolor{green!100} ! & \cellcolor{green!30} I & \cellcolor{green!8} 1 & y & \cellcolor{green!10} -- & \cellcolor{green!4} , & \cellcolor{green!2} by \\
m & \cellcolor{green!47} in & \cellcolor{green!4} my & \cellcolor{green!1} -- & z & \cellcolor{green!0} a & \cellcolor{green!0} ? & \cellcolor{green!0} \\

    \end{tabular}
    \caption{Top replacements for each lower case character - darker colors represent higher frequency of occurrence}
    \label{tab:char_occurence}
\end{table}

\section{Conclusion}

In this paper, we demonstrated how to improve the quality of an important corpus of digitized books, by correcting transcription errors that generally occur due to OCR. Our key idea to provide ground truth was to identify thousands of duplicate books (titles scanned in different locations and of uncertain quality). We aligned them at the token level to find where the differences occur, and used modern language models to determine which book copy is of higher quality. Additionally, we used this alignment as training data to train a model for correcting OCR errors in singleton books (books without any duplicates).

We showed that our methods correct over six times as many errors as it introduces, and also demonstrate that our errors tend to be semantically sensible. Through our efforts, we produced a substantially better version of over 50,000 distinct titles from the Hathitrust and Guttenberg as a foundation for future NLP research as well as show some interesting analysis from post-OCR processing.

\newpage

\bibliography{anthology,custom}

\begin{thebibliography}{32}
\expandafter\ifx\csname natexlab\endcsname\relax\def\natexlab#1{#1}\fi

\bibitem[{Abedjan et~al.(2016)Abedjan, Chu, Deng, Fernandez, Ilyas, Ouzzani,
  Papotti, Stonebraker, and Tang}]{abedjan2016detecting}
Ziawasch Abedjan, Xu~Chu, Dong Deng, Raul~Castro Fernandez, Ihab~F Ilyas,
  Mourad Ouzzani, Paolo Papotti, Michael Stonebraker, and Nan Tang. 2016.
\newblock Detecting data errors: Where are we and what needs to be done?
\newblock \emph{Proceedings of the VLDB Endowment}, 9(12):993--1004.

\bibitem[{Afli et~al.(2016)Afli, Barrault, and Schwenk}]{afli2016ocr}
Haithem Afli, Lo{\"\i}c Barrault, and Holger Schwenk. 2016.
\newblock Ocr error correction using statistical machine translation.
\newblock \emph{Int. J. Comput. Linguistics Appl.}, 7(1):175--191.

\bibitem[{Amrhein and Clematide(2018)}]{amrhein2018supervised}
Chantal Amrhein and Simon Clematide. 2018.
\newblock Supervised ocr error detection and correction using statistical and
  neural machine translation methods.
\newblock \emph{Journal for Language Technology and Computational Linguistics
  (JLCL)}, 33(1):49--76.

\bibitem[{Bassil and Alwani(2012)}]{bassil2012ocr}
Youssef Bassil and Mohammad Alwani. 2012.
\newblock Ocr post-processing error correction algorithm using google online
  spelling suggestion.
\newblock \emph{arXiv preprint arXiv:1204.0191}.

\bibitem[{Bengio et~al.(2003)Bengio, Ducharme, Vincent, and
  Janvin}]{bengio2003neural}
Yoshua Bengio, R{\'e}jean Ducharme, Pascal Vincent, and Christian Janvin. 2003.
\newblock A neural probabilistic language model.
\newblock \emph{The journal of machine learning research}, 3:1137--1155.

\bibitem[{Chiron et~al.(2017{\natexlab{a}})Chiron, Doucet, Coustaty, and
  Moreux}]{chiron2017icdar2017}
Guillaume Chiron, Antoine Doucet, Micka{\"e}l Coustaty, and Jean-Philippe
  Moreux. 2017{\natexlab{a}}.
\newblock Icdar2017 competition on post-ocr text correction.
\newblock In \emph{2017 14th IAPR International Conference on Document Analysis
  and Recognition (ICDAR)}, volume~1, pages 1423--1428. IEEE.

\bibitem[{Chiron et~al.(2017{\natexlab{b}})Chiron, Doucet, Coustaty, Visani,
  and Moreux}]{chiron2017impact}
Guillaume Chiron, Antoine Doucet, Micka{\"e}l Coustaty, Muriel Visani, and
  Jean-Philippe Moreux. 2017{\natexlab{b}}.
\newblock Impact of ocr errors on the use of digital libraries: towards a
  better access to information.
\newblock In \emph{2017 ACM/IEEE Joint Conference on Digital Libraries (JCDL)},
  pages 1--4. IEEE.

\bibitem[{Coustaty et~al.(2018)Coustaty, Doucet, Jatowt, Nguyen
  et~al.}]{coustaty2018adaptive}
Micka{\"e}l Coustaty, Antoine Doucet, Adam Jatowt, Nhu-Van Nguyen, et~al. 2018.
\newblock Adaptive edit-distance and regression approach for post-ocr text
  correction.
\newblock In \emph{International Conference on Asian Digital Libraries}, pages
  278--289. Springer.

\bibitem[{Devlin et~al.(2019)Devlin, Chang, Lee, and
  Toutanova}]{devlin2018bert}
Jacob Devlin, Ming-Wei Chang, Kenton Lee, and Kristina Toutanova. 2019.
\newblock \href {https://doi.org/10.18653/v1/N19-1423} {{BERT}: Pre-training of
  deep bidirectional transformers for language understanding}.
\newblock In \emph{Proceedings of the 2019 Conference of the North {A}merican
  Chapter of the Association for Computational Linguistics: Human Language
  Technologies, Volume 1 (Long and Short Papers)}, pages 4171--4186,
  Minneapolis, Minnesota. Association for Computational Linguistics.

\bibitem[{Evershed and Fitch(2014)}]{evershed2014correcting}
John Evershed and Kent Fitch. 2014.
\newblock Correcting noisy ocr: Context beats confusion.
\newblock In \emph{Proceedings of the First International Conference on Digital
  Access to Textual Cultural Heritage}, pages 45--51.

\bibitem[{Gutenberg(n.d.)}]{project_gutenberg}
Project Gutenberg. n.d.
\newblock \url{www.gutenberg.org}.
\newblock Accessed: April 2021.

\bibitem[{{HathiTrust Digital Library}()}]{hathitrust_dataset}
{HathiTrust Digital Library}.
\newblock \url{https://www.hathitrust.org/}.

\bibitem[{Jatowt et~al.(2019)Jatowt, Coustaty, Nguyen, Doucet
  et~al.}]{jatowt2019deep}
Adam Jatowt, Mickael Coustaty, Nhu-Van Nguyen, Antoine Doucet, et~al. 2019.
\newblock Deep statistical analysis of ocr errors for effective post-ocr
  processing.
\newblock In \emph{2019 ACM/IEEE Joint Conference on Digital Libraries (JCDL)},
  pages 29--38. IEEE.

\bibitem[{Kissos and Dershowitz(2016)}]{kissos2016ocr}
Ido Kissos and Nachum Dershowitz. 2016.
\newblock Ocr error correction using character correction and feature-based
  word classification.
\newblock In \emph{2016 12th IAPR Workshop on Document Analysis Systems (DAS)},
  pages 198--203. IEEE.

\bibitem[{Kolak et~al.(2003)Kolak, Byrne, and Resnik}]{kolak2003generative}
Okan Kolak, William Byrne, and Philip Resnik. 2003.
\newblock A generative probabilistic ocr model for nlp applications.
\newblock In \emph{Proceedings of the 2003 Human Language Technology Conference
  of the North American Chapter of the Association for Computational
  Linguistics}, pages 134--141.

\bibitem[{Kukich(1992)}]{kukich1992techniques}
Karen Kukich. 1992.
\newblock Techniques for automatically correcting words in text.
\newblock \emph{Acm Computing Surveys (CSUR)}, 24(4):377--439.

\bibitem[{Lew et~al.(2021)Lew, Agrawal, Sontag, and Mansinghka}]{lew2021pclean}
Alexander Lew, Monica Agrawal, David Sontag, and Vikash Mansinghka. 2021.
\newblock Pclean: Bayesian data cleaning at scale with domain-specific
  probabilistic programming.
\newblock In \emph{International Conference on Artificial Intelligence and
  Statistics}, pages 1927--1935. PMLR.

\bibitem[{Liu et~al.(2019)Liu, Ott, Goyal, Du, Joshi, Chen, Levy, Lewis,
  Zettlemoyer, and Stoyanov}]{liu2019roberta}
Yinhan Liu, Myle Ott, Naman Goyal, Jingfei Du, Mandar Joshi, Danqi Chen, Omer
  Levy, Mike Lewis, Luke Zettlemoyer, and Veselin Stoyanov. 2019.
\newblock Roberta: A robustly optimized bert pretraining approach.
\newblock \emph{arXiv preprint arXiv:1907.11692}.

\bibitem[{Llobet et~al.(2010)Llobet, Cerdan-Navarro, Perez-Cortes, and
  Arlandis}]{llobet2010ocr}
Rafael Llobet, Jose-Ramon Cerdan-Navarro, Juan-Carlos Perez-Cortes, and Joaquim
  Arlandis. 2010.
\newblock Ocr post-processing using weighted finite-state transducers.
\newblock In \emph{2010 20th International Conference on Pattern Recognition},
  pages 2021--2024. IEEE.

\bibitem[{Mayfield et~al.(2009)Mayfield, Neville, and
  Prabhakar}]{mayfield2009statistical}
Chris Mayfield, Jennifer Neville, and Sunil Prabhakar. 2009.
\newblock A statistical method for integrated data cleaning and imputation.

\bibitem[{Nguyen et~al.(2020)Nguyen, Jatowt, Nguyen, Coustaty, and
  Doucet}]{nguyen2020neural}
Thi Tuyet~Hai Nguyen, Adam Jatowt, Nhu-Van Nguyen, Mickael Coustaty, and
  Antoine Doucet. 2020.
\newblock Neural machine translation with bert for post-ocr error detection and
  correction.
\newblock In \emph{Proceedings of the ACM/IEEE Joint Conference on Digital
  Libraries in 2020}, pages 333--336.

\bibitem[{Pasula et~al.(2003)Pasula, Marthi, Milch, Russell, and
  Shpitser}]{pasula2003identity}
Hanna Pasula, Bhaskara Marthi, Brian Milch, Stuart~J Russell, and Ilya
  Shpitser. 2003.
\newblock Identity uncertainty and citation matching.
\newblock In \emph{Advances in neural information processing systems}, pages
  1425--1432.

\bibitem[{Radford et~al.(2019)Radford, Wu, Child, Luan, Amodei, and
  Sutskever}]{radford2019language}
Alec Radford, Jeffrey Wu, Rewon Child, David Luan, Dario Amodei, and Ilya
  Sutskever. 2019.
\newblock Language models are unsupervised multitask learners.
\newblock \emph{OpenAI blog}, 1(8):9.

\bibitem[{Raffel et~al.(2019)Raffel, Shazeer, Roberts, Lee, Narang, Matena,
  Zhou, Li, and Liu}]{raffel2019exploring}
Colin Raffel, Noam Shazeer, Adam Roberts, Katherine Lee, Sharan Narang, Michael
  Matena, Yanqi Zhou, Wei Li, and Peter~J Liu. 2019.
\newblock Exploring the limits of transfer learning with a unified text-to-text
  transformer.
\newblock \emph{arXiv preprint arXiv:1910.10683}.

\bibitem[{Rigaud et~al.(2019)Rigaud, Doucet, Coustaty, and
  Moreux}]{rigaud2019icdar}
Christophe Rigaud, Antoine Doucet, Micka{\"e}l Coustaty, and Jean-Philippe
  Moreux. 2019.
\newblock Icdar 2019 competition on post-ocr text correction.
\newblock In \emph{2019 International Conference on Document Analysis and
  Recognition (ICDAR)}, pages 1588--1593. IEEE.

\bibitem[{Salazar et~al.(2020)Salazar, Liang, Nguyen, and
  Kirchhoff}]{salazar2019masked}
Julian Salazar, Davis Liang, Toan~Q. Nguyen, and Katrin Kirchhoff. 2020.
\newblock \href {https://doi.org/10.18653/v1/2020.acl-main.240} {Masked
  language model scoring}.
\newblock In \emph{Proceedings of the 58th Annual Meeting of the Association
  for Computational Linguistics}, pages 2699--2712, Online. Association for
  Computational Linguistics.

\bibitem[{Schulz and Kuhn(2017)}]{schulz2017multi}
Sarah Schulz and Jonas Kuhn. 2017.
\newblock Multi-modular domain-tailored ocr post-correction.
\newblock In \emph{Proceedings of the 2017 Conference on Empirical Methods in
  Natural Language Processing}, pages 2716--2726.

\bibitem[{Taghva and Stofsky(2001)}]{taghva2001ocrspell}
Kazem Taghva and Eric Stofsky. 2001.
\newblock Ocrspell: an interactive spelling correction system for ocr errors in
  text.
\newblock \emph{International Journal on Document Analysis and Recognition},
  3(3):125--137.

\bibitem[{Tan et~al.(2020)Tan, Joty, Varshney, and Kan}]{tan2020mind}
Samson Tan, Shafiq Joty, Lav~R Varshney, and Min-Yen Kan. 2020.
\newblock Mind your inflections! improving nlp for non-standard english with
  base-inflection encoding.
\newblock \emph{arXiv preprint arXiv:2004.14870}.

\bibitem[{Tong and Evans(1996)}]{tong1996statistical}
Xiang Tong and David~A Evans. 1996.
\newblock A statistical approach to automatic ocr error correction in context.
\newblock In \emph{Fourth Workshop on Very Large Corpora}.

\bibitem[{Underwood(2016)}]{underwood_dataset}
Ted Underwood. 2016.
\newblock {Metadata for English-language Literature in HathiTrust Digital
  Library Beyond 1923}.
\newblock \url{https://github.com/tedunderwood/hathimetadata}.

\bibitem[{Vobl et~al.(2014)Vobl, Gotscharek, Reffle, Ringlstetter, and
  Schulz}]{vobl2014pocoto}
Thorsten Vobl, Annette Gotscharek, Uli Reffle, Christoph Ringlstetter, and
  Klaus~U Schulz. 2014.
\newblock Pocoto-an open source system for efficient interactive postcorrection
  of ocred historical texts.
\newblock In \emph{Proceedings of the First International Conference on Digital
  Access to Textual Cultural Heritage}, pages 57--61.

\end{thebibliography}
\bibliographystyle{acl_natbib}

\end{document}